# Contrast and visual saliency similarity-induced index for assessing image quality

Huizhen Jia[1], Lu Zhang[2], and Tonghan Wang[1]

[1] Jiangxi Engineering Laboratory on Radioactive Geoscience and Big Data Technology, East China University of Technology, Nanchang, 330013, Jiangxi, China

[2] Univ Rennes, INSA Rennes, CNRS, IETR - UMR 6164, F-35000 Rennes, France

Corresponding author: Tonghan Wang (e-mail: thwang_seu@ 163.com).

This work was supported partly by the Natural Science Foundation of China under Grant 61762004, the Science and Technology Project founded by the Education Department of Jiangxi Province (GJJ170455) and the Open Fund Project of Jiangxi Engineering Laboratory on Radioactive Geoscience and Big Data Technology under Grant JELRGBDT201702. This work was also supported in part by the Ph.D. Research Startup Foundation of East China University of Technology under Grants DHBK2016119 and DHBK2016120.

**ABSTRACT** Image quality that is consistent with human opinion is assessed by a perceptual image quality assessment (IQA) that defines/utilizes a computational model. A good model should take effectiveness and efficiency into consideration, but most of the previously proposed IQA models do not simultaneously consider these factors. Therefore, this study attempts to develop an effective and efficient IQA metric. Contrast is an inherent visual attribute that indicates image quality, and visual saliency (VS) is a quality that attracts the attention of human beings. The proposed model utilized these two features to characterize the image local quality. After obtaining the local contrast quality map and the global VS quality map, we added the weighted standard deviation of the previous two quality maps together to yield the final quality score. The experimental results for three benchmark databases (LIVE, TID2008, and CSIQ) demonstrated that our model performs the best in terms of a correlation with the human judgment of visual quality. Furthermore, compared with competing IQA models, this proposed model is more efficient.

## I. INTRODUCTION

Image quality assessment (IQA) plays a vital role in numerous applications, such as compression, image acquisition and transmission. The end receiver of any visual signal is a human being, but subjective IQA is often costly, slow, and difficult to integrate into real-time image-processing systems. Therefore, it is urgent to develop a perceptual method that correlates closely with the human visual system (HVS). According to whether there is a reference image, we can classify the objective IQA metrics as the following three types [1]: the full-reference (FR) metric, in which there is a pristine image for comparison; the reduced-reference (RR) metric, where only partial information concerning the pristine image is valid; and the no-reference (NR) metric, in which the pristine image is not available at all. We propose an FR-IQA method in this paper since the FR metric is widely utilized to assess image-processing algorithms.

FR methods have made much progress recently due to the considerable efforts being made. Traditional metrics such as the peak signal-to-noise ratio (PSNR) are the most widely used metrics in image processing. However, by failing to consider the properties of the HVS, these metrics do not correlate well with human opinion [2]. Thus, many IQA metrics have been developed based on the HVS. These kinds of representative models are [3] and [4], which use the sensitivity of the HVS to differentiate visual signals, such as contrast, luminance, frequency, and the interactions among these signals. The other kind of IQA model that utilizes the HVS has been adapted to extract structure information. The representative model was the structural similarity model (SSIM) [5], which has better performance than previous models. Various SSIM-induced metrics later arose [6]-[8]. In [6], the authors presented a multiscale SSIM that produced better performance than its previous one. In [7], the authors proposed a three-component SSIM that assigned different weights to edges, textures and smooth regions. Wang et al. improved the MS-SSIM with an information content weighted (IW)-SSIM index [8] by adopting a new pooling strategy. The information fidelity criterion (IFC) [9] and visual information fidelity (VIF) [10] considered the FR-IQA issues as the information fidelity problem according to information theory, and VIF is the developmental version of the IFC. The apparent distortion index was proposed by Larson and Chardler [11], in which HVS is regarded as operating two different strategies when evaluating the quality of high-quality images and low-quality images. The studies in [12] demonstrate that SSIM, MS-SSIM, and VIF perform better than the others. However, SSIM and MS-SSIM have the same defects that all positions are considered to have the same importance when yielding a quality score based on the local quality map. After images are composed into distinctive subbands and then assigned subbands with different weights when pooling, VIF assigns every position within each subband the same importance. It is noted that different locations of an image may have different perceptual meaning to our HVS; therefore, such pooling strategies must be improved. The so-called feature-similarity index (FSIM) [14], using the weighted average as the pooling strategy, was proposed based on the HVS perception of image quality according to its low-level features. The FSIM employs two features, namely, phase congruency and gradient magnitude, to produce local similarity maps, and the phase congruency map is also taken as the weightiness since it can reflect the degree of the perceptual importance of a local block to our HVS. Later, Zhang et al. proposed a visual saliency-induced metric (VSI) [15] based on the assumption that the visual saliency (VS) map of an image correlates highly with perceptual quality. Three components, namely, VS, gradient modulus and chrominance, are first computed by locally comparing the distorted image with the reference image through the similarity function. Then, the VS component is used as a weighting function to measure the importance of a local image region. The weighting strategy may improve IQA accuracy over the models with average pooling to some extent, but the process may be costly to compute the weights. Furthermore, this pooling strategy is likely to make nonlinear predictions with human judgments [16]. The image gradient is a popular feature in IQA since it can effectively capture the image local structures to which the HVS is highly sensitive. To this end, Xue et al. in his work, proposed the gradient magnitude similarity deviation (GMSD) index [16], where image gradient magnitude maps are computed, and then the standard deviations (SDs) of these maps are treated as the overall image quality scores. Considering that the contrast can reflect the change of luminance and the HVS usually has the characteristic of multiresolution, we proposed a multiscale contrast similarity deviation in [17]. Based on the above analysis, recently developed FR-IQA metrics utilize features in relation to the HVS or adopt a good pooling strategy to design IQA models. The goals of effectiveness and efficiency should be considered when designing IQA models; however, most previous IQA models do not simultaneously reach these two goals. Therefore, in this paper, we attempt to fill this need. Accordingly, we develop a model that uses contrast and VS that are closely related to the HVS, and we adopt the summation of the deviation-based pooling strategy. The experimental results demonstrate that, in comparison with previously examined state-of-the-art models, the proposed model is efficient and promising.

## II. RELATED WORK

This section presents works most related to our paper, including a brief review of the two features applied in IQA and the pooling strategy existing in FR-IQA.

### A. CONTRAST AND VS APPLIED IN IQA

The use of contrast and VS to design an IQA model is not new. SSIM [5] employed contrast as a part of its features (the other two are luminance and structure). The contrast reflects the change of luminance, while the standard deviation is the range for indicating the distortion severity of an image. Therefore, contrast is a distinctive visual attribute that indicates image quality. In fact, we can define "high quality" as proper contrast and little distortion. Contrast masking is a phenomenon in which the flaws of an image are masked locally by the other stimulations in the image. Considering the above analysis, we have proposed the multiscale contrast similarity deviation (MCSD) [17] which showed high correlation with human opinions in the experimental results on the benchmark databases.

VS, however, is another good feature of IQA since the HVS is quite sensitive to it. The salient regions of a visual scene are very important to the HVS, since human beings pay more attention to these regions. In [15], the authors thoroughly investigated VS in IQA and employed VS information as the main weighting function to pool the quality score. Efforts have been made to use the VS feature to enhance performance when designing IQA models in [18]. For these reasons, the proposed IQA model was designed by using the contrast and VS features to describe quality; it is noted that, however, VS is not used as a weighting function, similar to previous studies.

### B. THE POOLING STRATEGY PRESENT IN FR-IQA

After computing the feature similarity maps, a pooling strategy is needed to yield the quality score in FR-IQA metric. The simplest is average pooling, and it is the most widely utilized pooling strategy; i.e., all elements in the local quality map (LQM) are averaged to determine the overall quality prediction. Considering that different local areas may contribute to the entire quality of an image with different impacts, weighting strategies are thus also widely adopted. In contrast to average pooling, weighted pooling may gain overall quality prediction accuracy to some extent, but it may be costly when computing the weights. In [16], a deviation-based pooling strategy is used that achieves good prediction performance. However, this strategy may have good performance only when using one feature. Our paper utilizes a new pooling strategy—the summation of deviation-based pooling—in which the quality is computed as the summation of the standard deviation of these two features similarity maps. By this manner, it overcomes deviation-based pooling strategy errors when using only one feature, and the drawback that the VS commonly uses as a weighting function in designing the IQA models is addressed.

## III. PROPOSED IMAGE QUALITY ASSESSMENT METRIC

The proposed metric has the same two-step framework as most IQA models and is operated as follows. First, two similarity maps, namely, the local contrast similarity map and the global VS map, are generated. Then, the weighted SDs of the two similarity maps are added to yield the final quality score.

### A. LOCAL CONTRAST SIMILARITY MAP AND GLOBAL VS SIMILARITY MAP

Contrast has been defined by 0-0, and there are three types: Weber, Michelson and RMS contrast. The first type is usually used to measure the local contrast of a single target that is seen against a uniform background, while Michelson contrast is mainly used to measure the contrast of a periodic pattern. However, in complex images, these uniformity or periodicity conditions are not always satisfied. RMS contrast is preferred for natural stimuli and efficiency calculations. The experimental results in 0 show that RMS contrast with the subjective contrast of natural images has a better correlation than other contrasts. Therefore, for natural images, we adopted RMS contrast, which is also used by SSIM 0 and MCSD 0. RMS contrast is given as:

$$c = \left[\frac{1}{(N-1)}\sum_{i=1}^{N}(I_i - \bar{I})^2\right]^{1/2} \quad (1)$$

where $\bar{I}$ is the mean.

Contrast maps for the pristine image and the distorted one are computed by using formula (1) in a local manner. The local contrast map for the pristine image and distorted one is represented by $LC_r$ and $LC_d$, respectively. Then, the local contrast similarity (LCS) for the two images that are being compared is defined as:

$$LCS(r,d) = \frac{(2LC_r \cdot LC_d + c1)}{(LC_r^2 + LC_d^2 + c1)} \quad (2)$$

where $c1$ is a constant that increases stability, and $LC_r$ and $LC_d$ are calculated by local computation of the pristine image $r$ and the distortion image $d$, separately. For grayscale images, the contrast can be defined as the luminance difference that distinguishes an object. Proper contrast change is very important to image quality.

In this paper, we adopt a saliency map generator called the spectral residual method (SR) 0 that extracts the SR of the input image in the spectral domain, and spatial domain-based saliency maps are then generated. This method has a prominent advantage over other methods; namely, it has low computational complexity. To make the algorithm more efficient, the VS map for the proposed model was evaluated on the reduced resolutions and not on the original image scale with a manner similar to GMSD 0, and this generated VS map is global, not local. The VS similarity is given as:

$$GVSS(r,d) = \frac{2vs_r \cdot vs_d + c2}{vs_r^2 + vs_d^2 + c2} \quad (3)$$

where $c2$ is another positive constant, $vs_r$ and $vs_d$ are the VS maps of the pristine image $r$ and the corresponding distortion image $d$, respectively, and $GVSS$ is the global VS similarity map.

### B. SUMMATION OF DEVIATION-BASED POOLING

The pooling strategy is very important to FR-IQA. The mean and weighted mean are the two common pooling techniques in the literature. In contrast to average pooling, weighted pooling can increase overall quality prediction accuracy to a certain degree, but it may increase time complexity since the weights need more time to compute. The SD pooling that is proposed in 0 may reflect overall quality more accurately than the mean pooling for gradient magnitude similarity. When using only one feature to compute the LQM, the conclusion can be made that SD pooling could perform better than the nominal pooling method. When the LQM is obtained by using different features, the SD pooling is not suggested for application because the interactions among these features may cause the evaluation of the local image quality to deteriorate. Based on this consideration, the local contrast map and global VS map are generated, and then, the SD summation pooling is utilized to score the final quality. The proposed method is different than the VSI; in the VSI, VS is used as a weighting function. Using these methods, the proposed model yields excellent performance. The final quality score with SD pooling is computed after the generation of the local contrast similarity map and the global VS similarity map:

$$S = w_1 \cdot SD(LCS) + w_2 \cdot SD(GVSS) \quad (4)$$

subject to

$$w_1 + w_2 = 1 \quad (5)$$

where $w_1$ and $w_2$ are the weights that indicate the importance of the local contrast similarity map and the global VS similarity map, respectively, and:

$$SD(LCS) = \sqrt{\frac{1}{N}\sum_{i=1}^{N}(lcs(i) - LCSM)^2} \quad (6)$$

where

$$LCSM = \frac{1}{N}\sum_{i=1}^{N} lcs(i) \qquad (7)$$

$$SD(GVSS) = \sqrt{\frac{1}{N}\sum_{i=1}^{N}(GVSS(i) - GVSSM)^2} \qquad (8)$$

and

$$GVSSM = \frac{1}{N}\sum_{i=1}^{N} GVSS(i) \qquad (9)$$

Therefore, the procedure to calculate the proposed metric is illustrated in Figure 1.

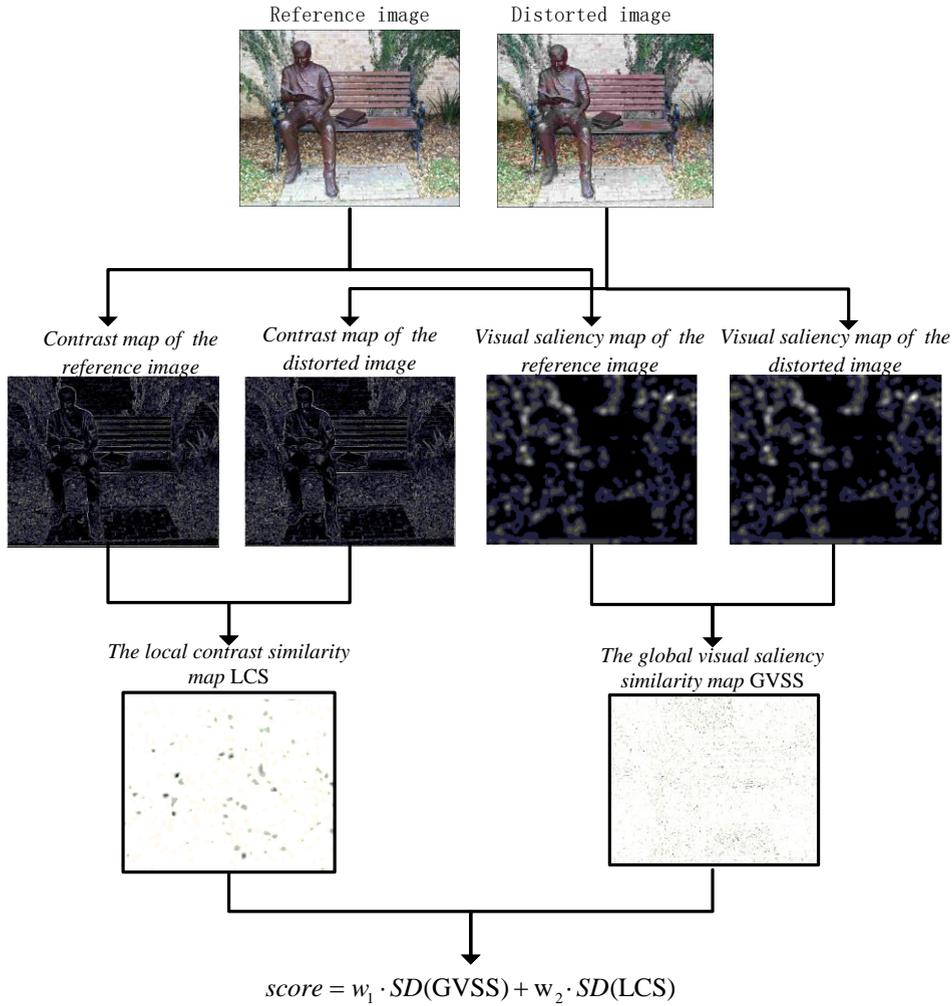

**FIGURE 1. Illustration of the proposed index computation**

To show the efficacy of this pooling strategy, the performance of different pooling strategies for the local contrast similarity map and global VS similarity map is shown in Figure 2. The "MEAN," "STD," and "MAD" are the mean, SD and mean absolute deviation 0, respectively, for the products of the two similarity maps. Meanwhile, the "Summation of mean-based pooling," the "Summation of MAD-based pooling," and the "Summation of deviation-based pooling" signify that we add the mean, mean absolute deviations and SDs of the two similarity maps together to obtain the quality score. Figure 2 shows that the proposed pooling strategy yields the best performance for the three benchmark databases of LIVE, TID2008, and CSIQ. More attention has been paid to the salient areas of a scene, which is consistent with SD pooling, since the SD is an indicator of the extent of distortion severities for an image. Furthermore, contrast is in the range of image luminance, and it can also be elaborately captured by the SD.

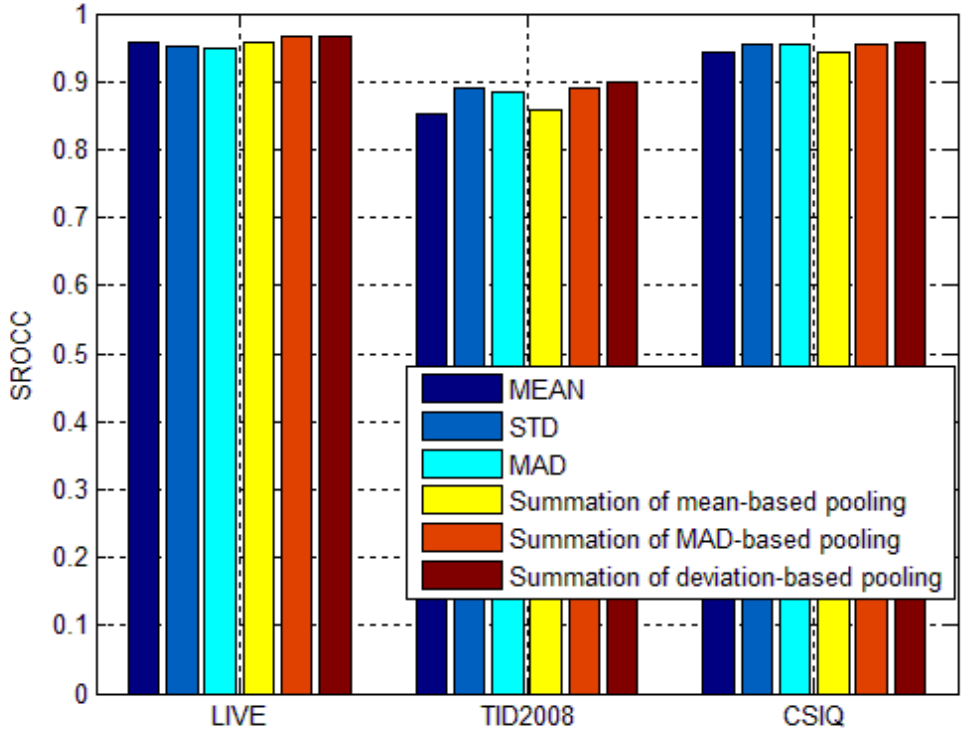

**Figure 2.** Performance of different pooling strategies

## IV. PERFORMANCE ESTIMATION

### A. ASSESSMENT CRITERIA AND DATABASES

To evaluate the effectiveness of our method, the three publicly available image databases were used for algorithm validation and comparison, including LIVE 0, TID2008 0 and CSIQ 0. Information about these three databases is summarized in Table I.

TABLE I.
BENCHMARK DATASET FOR EVALUATING IQA INDICES.

| Dataset | Reference Images | Distorted Images | Distortion Types | Observers |
|---|---|---|---|---|
| TID2008 | 25 | 1700 | 17 | 838 |
| CSIQ | 30 | 866 | 6 | 35 |
| LIVE | 29 | 779 | 5 | 161 |

Four commonly employed indices were calculated: Spearman's rank order correlation coefficient (SROCC); Kendall rank ordered correlation coefficient (KROCC) that estimates prediction monotonicity; Pearson's (linear) correlation coefficient (PLCC) that evaluates prediction linearity (regarded as the scale of prediction accuracy); and root mean squared error (RMSE) is used to measure prediction consistency. To compute the latter two indices, a logistic mapping is adopted to obtain the same scale values as subjective judgments 0.

$$p(x) = \alpha_1 \left( \frac{1}{2} - \frac{1}{1+e^{\alpha_2(x-\alpha_3)}} \right) + \alpha_4 x + \alpha_5 \quad (10)$$

where $\alpha_1$ is the parameter to be fitted, $x$ represents the original IQA scores, and $p(x)$ is the IQA score after the regression. $s$ is the subjective score, $d_i$ is the pairwise rank difference of $x$ and $s$, and $n$ indicates the element number of the dataset. Therefore, the following are used:

$$SROCC = 1 - \frac{6\sum_{i=1}^{n} d_i^2}{n(n^2-1)} \quad (11)$$

$$KROCC = \frac{n_c - n_d}{0.5n(n-1)} \quad (12)$$

where $n_c$ is the number of concordant pairs, and $n_d$ is the number of discordant pairs in the dataset. Let $(x_1, s_1), (x_2, s_2), \ldots (x_n, s_n)$ indicate a group of incorporated observations from the two stochastic variables of subjective judgments S and the scores X obtained by IQA metrics. For $(x_i, s_i)$ and $(x_j, s_j)$,

if both $x_i > x_j$ and $s_i > s_j$ or if $x_i < x_j$ and $s_i < s_j$, the observations are concordant:

$$PLCC = \frac{\overline{p}^T \overline{s}}{\sqrt{\overline{p}^T \overline{p} \; \overline{s}^T \overline{s}}} \quad (13)$$

where $\bar{p}$ and $\bar{s}$ are the mean-removed vectors of $p$ and $s$, given as:

$$RMSE = \sqrt{\frac{1}{n}\sum_{i=1}^{n}(s_i - p_i)^2} \qquad (14)$$

A value that approaches one for the former three indices demonstrates that the IQA metric has good performance. For the RMSE, a smaller value indicates better performance.

### B. PERFORMANCE COMPARISON

In the experiment, $c1 = 55$, $c2 = 0.00008$, $w_1 = 0.545$, and $w_2 = 0.455$, similar to the implementations of the SSIM 0, FSIM 0, and GMSD 0. Images $r$ and $d$ are first filtered by a $2 \times 2$ average filter and are then downsampled by a factor of 2.

The proposed method was compared with state-of-the-art methods, such as the SSIM 0, MS-SSIM 0, IW-SSIM 0, VIF 0, MAD 0, FSIM 0, GMSD 0, VSI 0 and MCSD 0. Table II tabulates the best three IQA models that are marked with boldface for the four indices.

TABLE II.
PERFORMANCE OF THE PROPOSED METRIC AND THE OTHER EIGHT COMPETING FR-IQA METRICS IN THREE BENCHMARK DATABASES. THE TOP THREE METRICS FOR EACH CRITERION ARE HIGHLIGHTED IN BOLDFACE.

|  |  | SSIM | MS-SSIM | IW-SSIM | VIF | MAD | FSIM | GMSD | VSI | MCSD | Proposed |
|---|---|---|---|---|---|---|---|---|---|---|---|
| TID2008 | SROCC | 0.7749 | 0.8542 | 0.8559 | 0.7491 | 0.8340 | 0.8804 | 0.8907 | **0.8979** | **0.8911** | **0.9001** |
|  | KROCC | 0.5768 | 0.6568 | 0.6636 | 0.5861 | 0.6445 | 0.6945 | 0.7092 | **0.7123** | **0.7133** | **0.7215** |
|  | PLCC | 0.7732 | 0.8451 | 0.8579 | 0.8084 | 0.8308 | 0.8738 | **0.8788** | 0.8762 | **0.8844** | **0.8961** |
|  | RMSE | 0.8511 | 0.7173 | 0.6895 | 0.7899 | 0.7468 | 0.6527 | **0.6404** | 0.6466 | **0.6263** | **0.5956** |
| CSIQ | SROCC | 0.8756 | 0.9133 | 0.9213 | 0.9195 | 0.9466 | 0.9310 | **0.9570** | 0.9423 | **0.9592** | **0.9580** |
|  | KROCC | 0.6907 | 0.7393 | 0.7529 | 0.7537 | 0.7970 | 0.7690 | **0.8122** | 0.7857 | **0.8171** | **0.8173** |
|  | PLCC | 0.8613 | 0.8991 | 0.9144 | 0.9277 | 0.9502 | 0.9192 | **0.9541** | 0.9279 | **0.9560** | **0.9589** |
|  | RMSE | 0.1334 | 0.1149 | 0.1063 | 0.0980 | **0.0818** | 0.1034 | **0.0786** | 0.0979 | **0.0770** | **0.0745** |
| LIVE | SROCC | 0.9479 | 0.9513 | 0.9567 | 0.9636 | **0.9672** | 0.9634 | 0.9603 | 0.9524 | **0.9668** | **0.9672** |
|  | KROCC | 0.7963 | 0.8045 | 0.8175 | 0.8282 | **0.8427** | **0.8337** | 0.8268 | 0.8058 | **0.8407** | 0.8406 |
|  | PLCC | 0.9449 | 0.9489 | 0.9522 | 0.9604 | **0.9688** | 0.9597 | 0.9603 | 0.9482 | **0.9675** | **0.9651** |
|  | RMSE | 8.9455 | 8.6188 | 8.3473 | 7.6137 | **6.7672** | 7.6780 | 7.6214 | 8.6816 | **6.9079** | **7.1573** |

Furthermore, according to Wang and Li's 0 suggestion, we provided the overall performance of the compared IQA models in Table III, where the weighted SROCC, KROCC, PLCC, and RMSE results are presented for the three databases. The weights are proportional to the total number of the distortion images each database has. The boldface font highlights the best performing model. The rankings are tabulated in Table IV.

Table III.
OVERALL PERFORMANCE OF THE IQA MODELS IN THREE DATABASES.

| IQA models | SROCC | KROCC | PLCC | RMSE |
|---|---|---|---|---|
| SSIM | 0.8413 | 0.6574 | 0.8360 | 2.5504 |
| MS-SSIM | 0.8921 | 0.7126 | 0.8833 | 2.4015 |
| IW-SSIM | 0.8963 | 0.7226 | 0.8945 | 2.3219 |
| VIF | 0.8432 | 0.6859 | 0.8747 | 2.1999 |
| MAD | 0.8942 | 0.7301 | 0.8939 | **1.9767** |
| FSIM | 0.9128 | 0.7462 | 0.9056 | 2.1466 |
| GMSD | 0.9241 | 0.7633 | 0.9173 | 2.1207 |
| VSI | 0.9221 | 0.7531 | 0.9064 | 2.3758 |
| MCSD | 0.9264 | 0.7698 | 0.9223 | **1.9470** |
| Proposed | **0.9307** | **0.7740** | **0.9284** | 1.9888 |

Table IV.
RANKING OF THE OVERALL PERFORMANCE OF THE IQA MODELS.

| IQA models | SROCC | KROCC | PLCC | RMSE |
|---|---|---|---|---|
| SSIM | 10 | 10 | 10 | 10 |
| MS-SSIM | 8 | 8 | 8 | 9 |
| IW-SSIM | 6 | 7 | 6 | 7 |
| VIF | 9 | 9 | 9 | 5 |
| MAD | 7 | 6 | 7 | 2 |
| FSIM | 5 | 5 | 5 | 6 |
| GMSD | 3 | 3 | 3 | 4 |
| VSI | 4 | 4 | 4 | 8 |
| MCSD | 2 | 2 | 2 | 1 |
| Proposed | **1** | **1** | **1** | **3** |

From the above tables, the proposed model clearly shows consistent good performance on all standard databases. Specifically, the proposed model had better performance than the other IQA metrics for the TID2008 and CSIQ databases. For LIVE, the proposed model performed only slightly worse than the best results from MAD. Notably, MAD works well for the LIVE database but fails to yield good results for the other two largest databases. The proposed model achieves the best performance in terms of the individual databases or the weighted average over the three benchmark databases. The proposed model is followed by the MCSD 0 and GMSD 0 and VSI 0. The performance of the VSI on the RMSE item is poor, whereas MCSD obtains the best performance for the RMSE.

## C. PERFORMANCE COMPARISONS ON DIFFERENT DISTORTION TYPES

A good IQA model should yield good performance overall and should also predict consistently well on each distortion type. To this end, Table V shows the results of competing IQA models with different distortion types. Because of space limitations and because the conclusions are similar with the use of the other measures (the KROCC, PLCC and RMSE), only the SROCC scores are shown. In Table V, we highlighted in boldface font to mark the top three IQA models for each distortion.

Table V.
SROCC values of the IQA model for each type of distortion.

|  |  | SSIM | MS-SSIM | IW-SSIM | VIF | MAD | FSIM | GMSD | VSI | Proposed |
|---|---|---|---|---|---|---|---|---|---|---|
| TID2008 | AGN | 0.8107 | 0.8086 | 0.7869 | 0.8804 | 0.8388 | 0.8574 | **0.9180** | **0.9229** | **0.9202** |
|  | ANC | 0.8029 | 0.8054 | 0.7920 | 0.8768 | 0.8258 | 0.8515 | **0.8977** | **0.9118** | **0.8972** |
|  | SCN | 0.8145 | 0.8209 | 0.7714 | 0.8709 | 0.8678 | 0.8485 | **0.9132** | **0.9296** | **0.9048** |
|  | MN | 0.7795 | **0.8107** | **0.8087** | **0.8683** | 0.7336 | 0.8023 | 0.7087 | 0.7734 | 0.7696 |
|  | HFN | 0.8729 | 0.8694 | 0.8662 | 0.9075 | 0.8864 | 0.9093 | **0.9189** | **0.9253** | **0.9184** |
|  | IN | 0.6732 | 0.6907 | 0.6465 | **0.8326** | 0.0650 | **0.7456** | 0.6611 | **0.8298** | 0.7006 |
|  | QN | 0.8531 | 0.8589 | 0.8177 | 0.7970 | 0.8160 | 0.8555 | **0.8875** | **0.8731** | **0.8883** |
|  | GB | **0.9544** | **0.9563** | **0.9636** | 0.9540 | 0.9197 | 0.9472 | 0.8968 | 0.9529 | 0.9304 |
|  | DEN | 0.9530 | 0.9582 | 0.9473 | 0.9161 | 0.9434 | 0.9604 | **0.9752** | **0.9693** | **0.9695** |
|  | JPEG | 0.9252 | 0.9322 | 0.9184 | 0.9168 | 0.9275 | 0.9282 | **0.9525** | **0.9616** | **0.9452** |
|  | JP2K | 0.9625 | 0.9700 | 0.9738 | 0.9709 | 0.9707 | 0.9775 | **0.9795** | **0.9848** | **0.9778** |
|  | JGTE | 0.8678 | 0.8681 | 0.8588 | 0.8583 | 0.8661 | **0.8708** | 0.8621 | **0.9160** | **0.8893** |
|  | J2TE | 0.8577 | 0.8606 | 0.8203 | 0.8501 | 0.8394 | 0.8542 | **0.8825** | **0.8942** | **0.8696** |
|  | NEPN | 0.7107 | 0.7377 | **0.7724** | 0.7619 | **0.8287** | 0.7494 | 0.7601 | **0.7699** | 0.7658 |
|  | Block | 0.8462 | 0.7546 | 0.7623 | 0.8324 | 0.7970 | **0.8489** | **0.8967** | 0.6295 | **0.8414** |
|  | MS | **0.7231** | **0.7338** | **0.7067** | 0.5102 | 0.5161 | 0.6695 | 0.6486 | 0.6714 | 0.6724 |
|  | CTC | 0.5246 | 0.6381 | 0.6301 | **0.8188** | 0.2723 | **0.6480** | 0.4659 | **0.6557** | 0.4662 |
| CSIQ | AGWN | 0.8974 | 0.9471 | 0.9380 | 0.9575 | 0.9541 | 0.9262 | **0.9676** | **0.9636** | **0.9670** |
|  | JPEG | 0.9546 | 0.9634 | **0.9662** | **0.9705** | 0.9615 | 0.9654 | 0.9651 | 0.9618 | **0.9689** |
|  | JP2K | 0.9606 | 0.9683 | 0.9683 | 0.9672 | **0.9752** | 0.9685 | **0.9717** | 0.9694 | **0.9777** |
|  | AGPN | 0.8922 | 0.9331 | 0.9059 | 0.9511 | **0.9570** | 0.9234 | 0.9502 | **0.9638** | **0.9516** |
|  | GB | 0.9609 | 0.9711 | **0.9782** | **0.9745** | 0.9602 | 0.9729 | 0.9712 | 0.9679 | **0.9789** |
|  | GCD | 0.7922 | **0.9526** | **0.9539** | 0.9345 | 0.9207 | 0.9420 | 0.9037 | **0.9504** | 0.9324 |
| LIVE | JP2K | 0.9614 | 0.9627 | 0.9649 | 0.9696 | 0.9692 | **0.9717** | **0.9711** | 0.9604 | **0.9719** |
|  | JPEG | 0.9764 | 0.9815 | 0.9808 | **0.9846** | 0.9786 | **0.9834** | 0.9782 | 0.9761 | **0.9836** |
|  | AWGN | 0.9694 | 0.9733 | 0.9667 | **0.9858** | **0.9873** | 0.9652 | 0.9737 | **0.9835** | 0.9809 |
|  | GB | 0.9517 | 0.9542 | **0.9720** | **0.9728** | 0.9510 | **0.9708** | 0.9567 | 0.9527 | 0.9662 |
|  | FF | 0.9556 | 0.9471 | 0.9442 | **0.9650** | **0.9589** | 0.9499 | 0.9416 | 0.9430 | **0.9592** |

The proposed model is among the top three models for 19 iterations, followed by VSI and GMSD, which are ranked among the top three models for 17 iterations and 13 iterations, respectively. Thus, the proposed model performed the best, while the VSI and GMSD demonstrated comparable performance when distortion types were specified. To visualize the competing IQA models' consistency on different distortion types, we drew scatter plots for the TID2008 database, as shown in Figure 3.

Figure 3 shows that the proposed model performs more consistently on different distortions than the competing models.

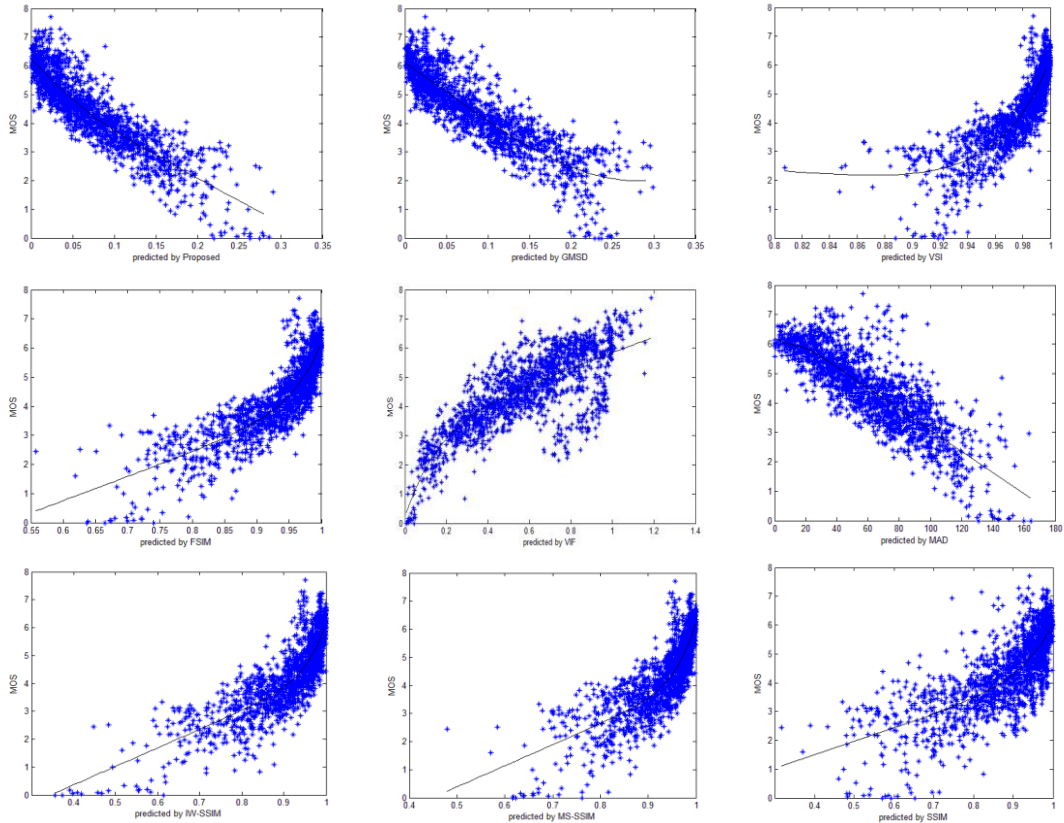

**Figure 3.** Scatter plots for the TID2008 database of the predicted quality scores when compared with the subjective quality scores (MOS) of the representative FR-IQA models

## D. COMPUTATIONAL COST

Efficiency, as the other crucial factor, should also be considered when designing a good IQA model. Therefore, experiments on run times were conducted. Table VI lists the amount of time (in seconds) needed to compute each quality measure on a color image with a resolution of $512 \times 512$ (taken from the CSIQ database) on a 2.66 GHz Intel Core2 Quad CPU with 5 GB of RAM, and boldface font was used to mark the top three models. All the IQA models (except the VSI because it is specially designed for color images) transformed the color image to a grayscale image.

Table VI.
Run times of the competing IQA models.

| IQA models | Running time (s) |
|---|---|
| SSIM | 0.0367 |
| MS-SSIM | 0.1643 |
| IW-SSIM | 0.8721 |
| VIF | 1.8774 |
| MAD | 2.7192 |
| FSIM | 0.5367 |
| GMSD | 0.0129 |
| VSI | 0.3044 |
| Proposed | 0.0443 |

Table VI shows that the GMSD, the SSIM and the proposed method were the top 3 most efficient IQA models and surpassed the other models by a large margin. For example, the proposed model is approximately 9 times faster than the VSI model, which can achieve state-of-the-art prediction performance.

## V. CONCLUSION

A good IQA model should be both effective and efficient, but most IQA models fail to satisfy these two criteria. Thus, in this paper, following these two criteria, we propose a new FR-IQA model based on the summation of a deviation-based pooling strategy for a local contrast similarity map and a global VS similarity map. We considered that contrast is an inherent attribute that can indicate image quality and that the VS map correlates highly with perceptual quality. Moreover, the proposed pooling strategy takes full advantage of these two features. Compared with the results for other competing state-of-the-art IQA models, the experimental results show that the proposed model performs better, making it an ideal candidate for IQA real-time applications. In addition, the proposed method can be improved with the emergence of more promising VS models.